\documentclass[letterpaper, 10 pt, conference]{ieeeconf}  

\IEEEoverridecommandlockouts                              

\usepackage{graphics} 
\usepackage{epsfig} 
\usepackage{mathptmx} 
\usepackage{amsmath} 
\usepackage[ruled,vlined]{algorithm2e}
\usepackage{graphicx}

\title{\LARGE \bf
Human Position Detection \& Tracking with On-robot Time--of--Flight Laser Ranging Sensors
}

\author{Sarthak Arora$^{1}$, Shitij Kumar$^{2}$, Ferat Sahin$^{3}$
\thanks{$^{1}$ Sarthak Arora *, Graduate Student,
        {\tt\small sa9472@rit.edu}}%
\thanks{$^{2}$ Shitij Kumar *, Ph.D. Candidate Engineering,
        {\tt\small spk4422@rit.edu}}%
\thanks{$^{3}$ Ferat Sahin *, Professor,
        {\tt\small feseee@rit.edu}}
    \thanks{* Department of Electrical and Microelectronics Engineering,
Rochester Institute of Technology, Rochester, NY 14623, USA}%
}

\begin{document}

\maketitle
\thispagestyle{empty}
\pagestyle{empty}

\begin{abstract}

In this paper, we propose a simple methodology to detect the partial pose of a human occupying the manipulator work-space using only on-robot time--of--flight laser ranging sensors. The sensors are affixed on each link of the robot in a circular array fashion where each array possesses sixteen single unit laser ranging lidar(s). The detection is performed by leveraging an artificial neural network which takes a highly sparse 3-D point cloud input to produce an estimate of the partial pose which is the ground projection frame of the human footprint. We also present a particle filter based approach to the tracking problem when the input data is unreliable. Ultimately, the simulation results are presented and analyzed. 

\end{abstract}

\section{INTRODUCTION}

With the onset of human robot collaboration (HRC), the interaction between the operator and the robot have become extremely human-centric. For any interaction to safely occur, information associated with human/operator position with-respect-to the robot must be present. Usually, these scenarios are rife in factory floors and indoor environments. Therefore, the use of exteroceptive sensor systems such as \cite{optitrack} and \cite{vicon}
have enabled complete human tracking including bio-mechanical information. It must be noted that these sensing systems are setup and mounted in the robot's environment and usually require calibration routines and planning of sensor placement around the concerned volume of operation. However, due to the densely occluded nature of indoor environments and factory floors, occlusion becomes inevitable. To alleviate this problem the use of exteroceptive sensors affixed to the robot is a viable option.

As it can be verbose and confusing to refer to the aforementioned sensors with their designated terms. For convenience, the systems can be divided into two categories similar to virtual-reality (VR) tracking systems. When the tracking system is completely self-contained within the VR headset it is referred to as "inside-out" tracking. When the tracking system is completely external to the VR headset it is referred to as "outside-in" tracking. Similarly, when a sensing system is affixed on the robot it would be convenient to express the system as an "inside-out" sensing system from the robot's perspective and "outside-in" from the sensors mounted in the environment. A similar idea was proposed in \cite{dynamic} where the author(s) classified "inside-out" \& "outside-in" as intrinsic \& extrinsic sensing systems respectively. They also laid out the ground work for the sensing system demonstrated in this work. In \cite{tof-ssm}, several arguments were presented to demonstrate the similarity between the simulated and the physical versions of the time--of-flight laser ranging sensing system. An ISO compliant \cite{iso} safety algorithm was also presented in \cite{tof-ssm}, however, the approach only leveraged distance thresholds to design the controller. The motivation of this work is to augment the work done in \cite{tof-ssm} with a human position estimation system using solely the "inside-out" sensing system.

\begin{figure}
  \includegraphics[width=\linewidth]{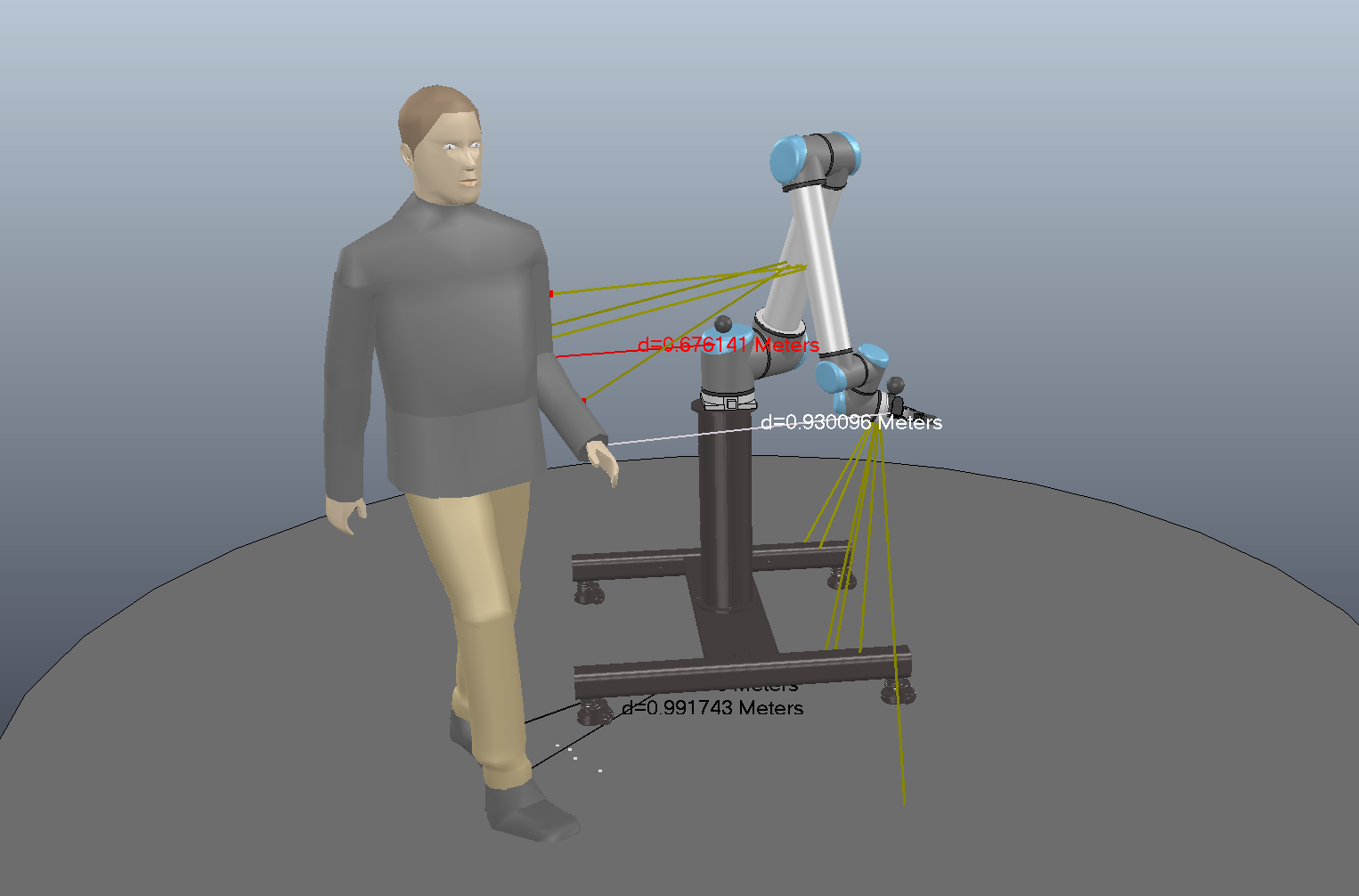}
  \caption{UR10 robot with time--of--flight sensors mounted on each link, rays are shown when hit. The observation from each lidar are used to estimate and track the human operator in the robot workspace.}
  \label{fig:cover}
\end{figure}
Our approach is inspired by several works done in the field of object detection and vehicle tracking. In \cite{pfcar}, the authors present a particle filter based approach to track a single vehicle that passes by using six ultrasonic sensors distributed around the vehicle. In \cite{kalman}, the authors design an Extended Kalman Filter using 8 ultrasonic sensors mounted on each side of the vehicle to track other objects. It should be noted that aforementioned approaches assume the vehicles to be completely on the ground and therefore, track the ground projection(s) of the objects. Levying this assumption, the authors of \cite{pixor}, use 3-D point clouds to generate the top view of the point clouds and slice the cloud along the z-axis to get 2-D slices of the points. In \cite{multi-view}, the authors use the top view of the point cloud to pass it to a deep learning model \cite{deep}.

As our environment is indoors and controlled, our problem formulation is greatly simplified as compared to the approaches mentioned above. It is assumed that the robot workspace is sporadically occupied by one person only and the 3-D point clouds generated are down-projected to 2D point maps for denser inputs with limited pose information. Ultimately, our work holds the following contributions:
\begin{itemize}
    \item A simple approach to human position estimation and tracking that solely uses "inside-out" sensing systems. 
    \item With work done in \cite{tof-ssm}, the work presented in this paper can be used as a perception strategy for the safety controller.
\end{itemize}

\section{Problem Formulation}

\subsection{Setup \& Assumptions}

The system presented in this work comprises of three circular arrays which consist sixteen single unit lidar(s) each with a field of view of 25 degrees on every single unit lidar. The pose of the human occupying the workspace is considered to be a two dimensional partial pose due to the limited sensing capability of the system.

The human pose is represented by the 2-D location of the human on the workspace floor. The pose is computed by only considering ${x_{human}}$ \& ${y_{human}}$ with respect to a fixed reference frame. Therefore, it can be represented as: 
\begin{equation}
    {}^{World} T_{Human} = 
\begin{bmatrix}
    x_{\emph{h}} \\
    y_{\emph{h}} \\
    0
  \end{bmatrix}
\end{equation}

Where, ``$\emph{World}$" represents the fixed reference frame in the center of the workspace. It should be noted that the third coordinate (${z-axis}$) has been discarded.

It should be noted that the robot's base frame is created by projecting it directly on top of the world frame.  
As each sensor array is circular, it is mounted like a ring around each link of the robot such that there is zero translation and rotation between the ring center and link center. This enables the sensors to be a part of the robot's kinematic chain. Therefore, the distance reported by each single unit lidar can be converted a 3-D point observed with respect to the world frame. Each lidar unit can observe distance of upto 2m. This can be imagined as a distributed lidar spread across the robot's body surface. Each distance transformation reported by each lidar in each ring can be reported as:
\begin{subequations}
\begin{equation*}
 {}^{World}T_{{Lidar\hspace{2pt}Observation}_{i,j}}=  
\end{equation*}
\begin{equation}
{}^{World}T_{{Robot Footprint}} \\ \bullet ^{Ring_i}T_{{Lidar_j}}\\ \bullet {}^{Lidar_j}T_{{Observation}}
\end{equation}
\end{subequations}
Where ${i=1,2,3}$ \& ${j=1,2,...,16}$ represent the ring and the lidar index respectively. \\\\ 
The lidar observation returned is a 3D point from the above mentioned kinematic-chain. However, as the lidar reports any observed distance, association of that data with a human subject can be difficult. 
\subsection{Data Association}
In \cite{tof-ssm}, a method for self-occlusion detection is shown where a physics engine is used for data association. Only the lidar observations that are not associated with the robot are kept, as there is only one human occupying the workspace, it is safe to say that the filtered out observations are therefore associated with the human.

\section{Methodology}

The lidar observations from each ring that are associated with the human are collected along with the human pose with-respect-to the ``${World}$" frame. The collected data is then used to develop a test-train split in the dataset for training an artificial neural network. The output of the artificial neural network is then used as input to a particle filter that provides the tracking capability to the system. 

\subsection{Data Collection \& Processing} 
The data is collected by time synchronization of the entire system. As a simulator is used for the experiment, each lidar observation, joint angles of the robot and the human pose is hard time synced with each other for getting the complete state of the state at any given time.

\subsection{Artificial Neural Network}

The artificial neural network learns the mapping between the lidar observations \& joint angles of the robot with respect to the human pose. The network receives an input vector of size 54. Where:
\begin{equation}
    {Input\hspace{2pt}Vector, z_n} = 
    \begin{bmatrix}
        Pose_{{Lidar\hspace{2pt}Obs}_{1,1,...,16}} \\
        Pose_{{Lidar\hspace{2pt}Obs}_{2,1,...,16}} \\
        Pose_{{Lidar\hspace{2pt}Obs}_{3,1,...,16}} \\
        {\theta}_{1,...,6}
    \end{bmatrix}
    ;n = 1,...,N
\end{equation}

Each element in ${x_n}$ is vertical vector. The ${Pose_{{Lidar\hspace{2pt}Obs}_{i,j}}}$ vector is a vector of size ${16}$x${1}$ where each element is the 3D observation reported by a lidar ${j}$ from ring ${i}$. The ${\theta_q}$ is a vector of the robot's joint angles of size $6$x$1$.

The output of the network is the human pose which essentially a vector of size $2$x$1$. The network is essentially a regressor that takes the input vector and outputs a continuous value where the ground truth is the human pose recorded during the simulation. Therefore, the network output is characterized by:

\begin{equation}
    {y}_n = 
        \begin{bmatrix}
        x_h \\
        y_h
    \end{bmatrix}
    ;n = 1,...,N
\end{equation}
Where n is the size of the training data.

The network essentially act like a detector which regresses over inputs to produce a continuous pose. It can be formulated in a parameterized form:

\begin{equation}
    y_n = f(z_n, W,B)
\end{equation}

\subsection{Network Training \& Architecture}
The network has a simple structure and comprises of three hidden layers with Relu \cite{relu} and tanh non-linear activations. The output layer of the network is identity or it directly produces the logits in the output. The network also possesses a dropout \cite{dropout} layer before the output layer to avoid overfitting. \\

The network loss function was modified to be Root Mean Square Error as the data is essentially pose data, therefore, we attempt to directly learn a mapping. The loss is given by:

\begin{equation}
    \mathcal{L} = \sqrt{\dfrac{1}{n}\sum_{n=1}^{N} (y_n - {}^{true}y_n)^2}
\end{equation}

The input data for the network was subjected to gaussian noise as the simulator provides ideal condition to simulate sensor noise. To make the network robust to estimating the human pose, some gaussian noise was also injected in the ground truth and data augmentation was done.

\subsection{Tracking with Particle Filter}

The output of neural network was used as input to a particle filter where each particle is represented by a state vector:

\begin{equation}
    X_t =  
    \begin{bmatrix}
        \widehat{x_h} \\
        \widehat{y_h} \\
        \dot{x_h} \\
        \dot{y_h} \\
    \end{bmatrix}
\end{equation}

The state vector was subjected to input noise as shown in \cite{pfcar}. The particle filter algorithm was implemented as done in \cite{pftut}. At first, it seems obvious to directly fuse the reading from each single unit lidar in the system, however, the single unit lidar(s) are observing only a few different points in the human body at a time. The sensor independence would lead to biased estimation the raw observations were passed to the filter. To overcome the sparsity in the observations the joint angles of the robot were added to the neural network inputs. 
\begin{algorithm}
\SetAlgoLined
\KwResult{$X_t$}
 initialize particles\;
 initialize neuralnet\;
 initialize timestep\;
 \While{true}{

  \eIf{inputIsValid()}{
   output = net(input)\;
   prediction = predict(particles, timestep)\;
   correction = correct(particles, output)
   }{
   prediction = predict(particles, timestep)\;
  }
 }
 \caption{Estimation and Tracking Algorithm}
\end{algorithm}

Shown above, is the algorithm being used, it should be noted that the algorithm first checks for valid input and then moves on to start making predictions. An invalid input is obtained when no input observations are reported, therefore the algorithm then moves on to making predictions using the motion model which in this case is a fixed velocity model.

\section{Simulation Experiment Results}
The experiment was performed by randomizing the robot and the human trajectories. In other words, the experiment was run for an arbitrary amount of time while the data was collected from the simulator. The joint angles of the robot were rotated by drawing angle values from a uniform distribution and then passing those values to a joint controller. On the other hand, the human trajectory was randomized by inducing random turn in the walking path of the human.

\begin{figure}
  \includegraphics[width=\linewidth]{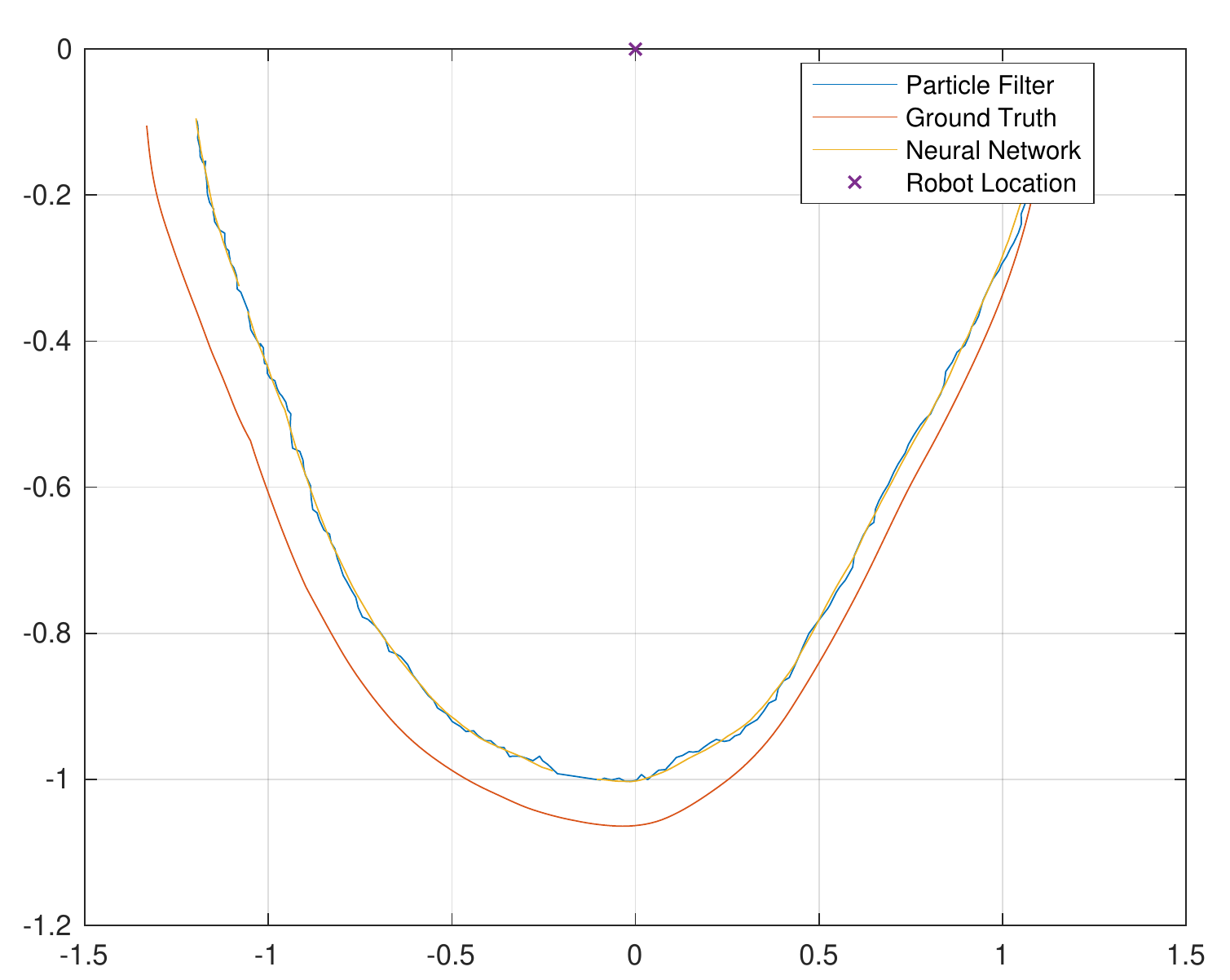}
  \caption{A figure showing the trajectory of the human around the robot (located at (0,0)). It can be seen from the figure that ground truth is biased away from the neural network and the particle filter plots. That can be explained by the fact that the true human pose is always further away from the outermost points on the human body surface as observed by the lidar(s).}
  \label{fig:boat1}
\end{figure}

\begin{table}[h]
\centering
\begin{tabular}{l|l|l|}
\cline{2-3}
                           & Neural Net & Particle Filter \\ \hline
\multicolumn{1}{|l|}{RMSE} & 0.1196                     & 0.1061                          \\ \hline
\end{tabular}
\end{table}

After obtaining the RMSE values, it can be seen that the particle filter has a lower RMSE of 0.1061 as compared to the Neural Net which is due to the innovation done by the particle filter when no inputs were present, it can be thought of as imputing the values when none are present. 
It can be seen from the results that the bias parameters $B$ in the neural network are essentially responsible for controlling the offset between the ground truth trajectory and the predicted trajectory. On the other hand, the weights $W$ are responsible for controlling the significance of each lidar observation. 
The particle filter performance could be tuned primarily by changing the number of particles it requires, for higher quantities the particle filter generated a smoother trajectory, however increasing the particles can greatly increase the computation time.

The neural network performance was greatly affected by introducing a tanh activation in the dense layers of the network. The network was trained using Stochastic Gradient Descent \cite{sgd} with nesterov momentum \cite{momentum}. A decaying learning rate was used with a factor of 0.000001 where the learning rate was kept around 0.01. It was also observed that introduction of RMSE as a loss function made it penalize the network much higher than mean squared error. 

\section{Conclusion and Future Work}

As it can be seen from the simulation results that using a neural network as an estimator for sparse point clouds is a possible, it can be safely said that using more powerful methods such as convolutional neural network (CNNs) \cite{cnn}. Using techniques to project point cloud information into occupancy grids \cite{thrun} can potentially create an opportunity for using CNNs. The approach can also benefit from applying sensor fusion directly to the lidar observations, however, it was noted that using joint angles of the robot along with the observations greatly enhanced the network performance because the joint angles act as a context vector in the network and were effective when only one or more 3D point was available in the point cloud.

\bibliographystyle{IEEEtran}
\bibliography{my_bib2}{}

\end{document}